\documentclass{article}
\pdfoutput=1
\usepackage{amsmath,epsfig}
\usepackage[preprint]{spconfa4}
\usepackage{xcolor}
\usepackage{cite}
\usepackage{graphicx}
\usepackage{amssymb}
\usepackage{booktabs}
\usepackage{multirow}
\usepackage{marvosym}
\usepackage{bbm}
\let\OLDthebibliography\thebibliography
\renewcommand\thebibliography[1]{
  \OLDthebibliography{#1}
  \setlength{\parskip}{0pt}
  \setlength{\itemsep}{0pt plus 0.3ex}
}

\begin{document}\sloppy

\def\x{{\mathbf x}}
\def\L{{\cal L}}

\title{HDNet: A Hierarchically Decoupled Network for Crowd Counting}
%
\name{Chenliang Gu, Changan Wang, Bin-Bin Gao, Jun Liu, Tianliang Zhang\textsuperscript{\Letter}}

\address{Tencent YouTu Lab, Shenzhen, China \\
         \tt\small{\{guchenliang1996, wangchangan.hust, csgaobb, junsenselee, tianliangjay\}@gmail.com}
         }
\maketitle

\begin{abstract}
Recently, density map regression-based methods have dominated in crowd counting owing to their excellent fitting ability on density distribution. However, further improvement tends to saturate mainly because of the confusing background noise and the large density variation. In this paper, we propose a Hierarchically Decoupled Network (HDNet) to solve the above two problems within a unified framework. Specifically, a background classification sub-task is decomposed from the density map prediction task, which is then assigned to a Density Decoupling Module (DDM) to exploit its highly discriminative ability. For the remaining foreground prediction sub-task, it is further hierarchically decomposed to several density-specific sub-tasks by the DDM, which are then solved by the regression-based experts in a Foreground Density Estimation Module (FDEM). Although the proposed strategy effectively reduces the hypothesis space so as to relieve the optimization for those task-specific experts, the high correlation of these sub-tasks are ignored. Therefore, we introduce three types of interaction strategies to unify the whole framework, which are Feature Interaction, Gradient Interaction, and Scale Interaction. Integrated with the above spirits, HDNet achieves state-of-the-art performance on several popular counting benchmarks.
\end{abstract}
\begin{keywords}
Crowd Counting, Density Decoupling, Foreground Density Estimation, Interaction
\end{keywords}

\vspace{-0.6em}
\section{Introduction}
\label{sec:intro}

Crowd counting aims to estimate the number of persons in a still image or video frame, which recently draws increasing attention in the application of public safety. 

Despite the progressive advancement in crowd counting, there are still two tricky problems remaining to be solved: the cluttered background noise \cite{Modolo2021} and the large density variation \cite{song2021choose}. For the former, some methods \cite{miao2020shallow} focus on learning more robust features, but ignore that it is intrinsically hard to directly regress an exact \textit{zero} value for various background regions. Instead, some other methods \cite{Modolo2021} introduce a segmentation branch to exploit the highly discriminative ability of classification network. But they treat foreground with various densities as a single class, which ignores the large intra-class variation within foreground regions and leads to inferior background modeling accuracy. For the latter problem, \cite{song2021choose} adopts a multi-column based feature fusion strategy without taking the corresponding relations between receptive field and density distribution into consideration, thus leading to significant feature redundancy. Differently, we bypass these defects by simultaneously solving the two problems with a novel hierarchically decoupled strategy under a unified framework, being more simple, effective and consistent.


We firstly decompose a background classification sub-task from the whole density prediction task, inspired by its highly discriminative ability. Then, according to the spatial density distribution, the remaining foreground prediction sub-task is further hierarchically decomposed to several density-specific sub-tasks. The foreground related sub-tasks are solved by several regression based experts in a Foreground Density Estimation Module (FDEM). To guide the above decoupling process, we propose a Density Decoupling Module (DDM) which is supervised by a fine-grained classification loss. The proposed hierarchically decoupled strategy helps each task-specific expert to focus on their most skilled sub-task, collaboratively contributing to the final prediction. In addition, three types of interaction strategies, including Feature Interaction, Gradient Interaction and Scale Interaction, are introduced for the use of the intrinsic relations among these sub-tasks. Combining with the above spirits, we propose an effective and compact counting model termed as Hierarchically Decoupled Network (HDNet), achieving state-of-the-art performance on several popular benchmarks.


Contributions of our paper are summarized as follows:
\begin{itemize}
\item We propose a novel hierarchically decoupled strategy for crowd counting to simultaneously solve its two tricky problems, \textit{i.e.}, the cluttered background noise and the large density variation.
\item We propose three types of interaction strategies to collaboratively integrate the decoupled components, yielding a compact and unified counting model.
\item We demonstrate the effectiveness of our contributions through sufficient experiments, setting new performance records on several popular counting benchmarks. 
\end{itemize}

\section{Related Work}

Recently, Convolutional Neural Networks (CNN) based counting methods have achieved promising improvements. In this section, we review representative methods which focus on the following two problems: the cluttered background noise and the large density variation. 

\begin{figure*}[th]
\centering
\includegraphics[width=1.0\textwidth]{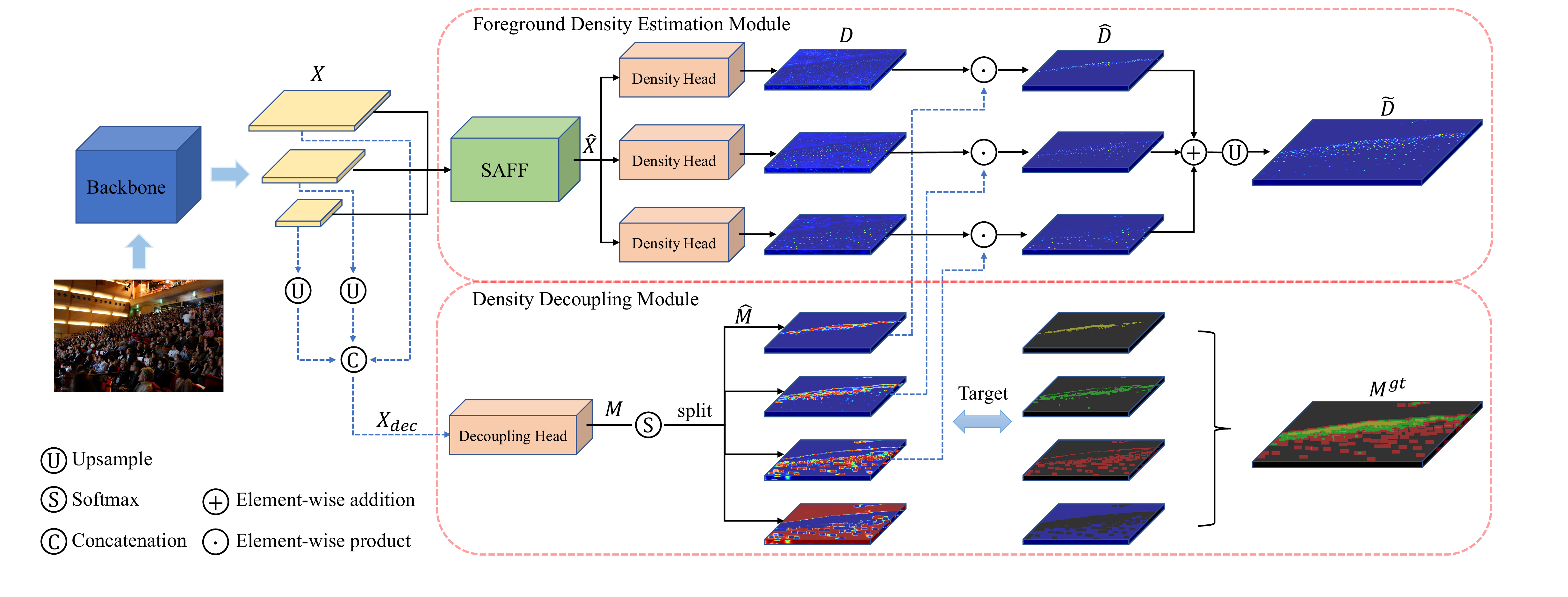} 
\caption{The overall architecture of the proposed Hierarchically Decoupled Network mainly consists of two components: Density Decoupling Module and Foreground Density Estimation Module. Scale Adaptive Feature Fusion (SAFF) block is introduced to fuse high-resolution and rich context information.}
\vspace{-1.0em} 
\label{fig2_framework}
\end{figure*}

\subsection{Background-robust Crowd Counting.}
The impact from cluttered background noise attracts increasing attention in the literature but remains a tricky problem. Specifically, \cite{arteta2016counting} introduces an explicit foreground-background segmentation to its multi-task architecture. Then the predicted foreground masks are used to form a better learning target. \cite{wan2019residual} trains a semantic prior network on the ADE20K dataset to reweight the feature maps for crowd counting, and the semantic prior helps to eliminate the side effect of noisy false alarms in background region. For the first time, \cite{Modolo2021} shows the importance of tackling with the noisy background problem by quantifying the counting errors from background region. Then they introduce a simple foreground segmentation branch to suppress background mistakes. However, previous works ignore the large intra-class variance within foreground regions and result in inferior background modeling accuracy. In this paper, we propose a simple and unified framework to deal with this problem, in which a fine-grained supervision for foreground helps the accurate modeling for background region. 

\subsection{Density-aware Crowd Counting.}
The major density variation problem is a long-standing challenge in crowd counting. To remedy this issue, some methods 
\cite{Zhang2016, Li2018} try to fuse multiple columns convolutions with different kernel sizes or receptive fields to obtain density-invariant features. Despite their effectiveness, they ignore the corresponding relations between receptive fields and density distributions, thus leading to significant feature redundancy \cite{song2021choose}. Besides, attention mechanism is also exploited to tackle with the density variation problem. 
Specially, DADNet \cite{Guo2019} uses different dilated rates in each parallel column to obtain attention maps and multi-scale features. ADCrowdNet \cite{liu2019adcrowdnet} designs an attention map generator to indicate the locations of congested regions for latter density map estimator. 
ASNet \cite{Jiang2020} learns attention scaling factors and automatically adjusts the density regions by multiplying density attention masks on them. These sub-tasks are optimized separately, which ignores the correlation between tasks.
Differently, with the proposed hierarchically decoupled strategy, we distribute regions with various densities to multiple density-specific experts to collaboratively contribute to the final prediction, accounting for the internal relations between receptive field and density distribution whilst being more simpler.

\section{The Proposed Approach}

In this section, we first describe the Hierarchically Decoupled Network, which is combined with the Density Decoupling Module and the Foreground Density Estimation Module. Then, we introduce the three interaction strategies including Feature Interaction, Scale Interaction, and Gradient Interaction.

\subsection{Density Decoupling Module}

As shown in Fig. \ref{fig2_framework}, considering that feature maps on different levels have various resolutions, we firstly construct the feature $\mathbf{X}_{dec}$ in decoupling branch with multi-level feature maps $\{\mathbf{X}_i\}_{i=1}^{n}$ from the backbone via upsampling.
The decoupling head of our DDM is simply equipped with a 3$\times$3 convolution block followed by a 1$\times$1 convolution block. The convolution block consists of a convolution layer, a batch normalization layer and ReLU layer. Each decoupling head converts $\mathbf{X}_{dec}$ into $\mathbf{M}\in\mathbb{R}^{\frac{H}{4} \times\frac{W}{4}\times(n+1)}$, where $n$ is the number of density levels, plus 1 for background. Note that, DDM only decouples foreground and background if we set $n$ to 1. When $n>1$, DDM is responsible not only for decoupling foreground and background but also for decoupling foreground into regions with multiple densities.

Accounting for the continuous spatial density variation across an image, we limit the range of activation values in $\mathbf{M}$ to $[0, 1]$. To this end, $\mathbf{M}$ is fed into a softmax function to get its soft output $\mathbf{\hat M}=\{\mathbf{\hat M}_i\}_{i=0}^n$, that is 
\begin{equation}
\mathbf{\hat M}_i^{j,k}=\frac{e^{\mathbf{M}_{i}^{j, k}}}{\sum_{i=0}^{n} e^{\mathbf{M}^{j, k}_{i}}},
\end{equation}
the value of $\mathbf{\hat M}_i^{j,k}$ represents a probability that it belongs to the $i$-th density level at spatial location $(j,k)$. The soft output $\mathbf{\hat M}$ is supervised by the ground-truth $\mathbf{M}^{gt}$ as following:
\begin{equation}
\mathcal{L}_{dec}=\mathcal{L}(\mathbf{\hat M}, \mathbf{M}^{gt}), \label{loss}
\end{equation}
where $\mathcal{L}(\cdot)$ denotes the cross-entropy loss. The ground-truth $\mathbf{M}^{gt}$ is generated in the similar way as \cite{Hossain2019}.

\vspace{-0.6em}
\subsection{Foreground Density Estimation Module}

We propose a Scale Adaptive Feature Fusion (SAFF) block to get multi-scale features$\{\mathbf{\hat X}_i\}_{i=1}^{n}$, which will be described in detail in later section. 

The Foreground Density Estimation Module (FDEM) contains $n$ density heads, and each density head is responsible for predicting image regions within a specific range of density. Similar to the architecture of the decoupling head, our each density head is implemented by a 3$\times$3 convolution block followed by a 1$\times$1 convolution block. The $i$-th density head takes $\mathbf{\hat X}_i$ as input and predicts the corresponding single-channel density map to 
produce $ \{\mathbf{D}_i\}_{i=1}^{n}$.
Then, in order to weaken the learning of the density prediction of the background region and focus on the different density regions of the foreground, we take the $n$ foreground soft masks $\{\mathbf{\hat M}_i\}_{i=1}^{n}$ as the attention maps to multiply the corresponding foreground density map $\mathbf{D}_i$. Then the final density map $\mathbf{\widetilde{D}}$ is the sum of the prediction results of all density heads, and it is calculated as : \begin{equation}
 \mathbf{\widetilde{D}}=\sum_{i=1}^{n} \mathbf{\hat D}_i =\sum_{i=1}^{n} \mathbf{D}_i \odot \mathbf{\hat M}_i, 
\end{equation}
where $\odot$ denotes the Hadamard operation. 

The overall loss of the hierarchically decoupled framework is defined as 
\begin{equation}
\mathcal{L} = \mathcal{L}_{reg} + \lambda \mathcal{L}_{dec},
\end{equation}
where ${L}_{reg}$ is $L_2$ loss between predicted density map and ground-truth density map, and $\lambda$ is a weight which balances the importance between $\mathcal{L}_{reg}$ and $\mathcal{L}_{dec}$.

\vspace{-0.6em}
\subsection{Interaction in the HDNet}
Density estimation and density decoupling are two highly correlated tasks in nature. 
Therefore, three types of interaction strategies are proposed to integrate these sub-tasks in consideration of the intrinsic relations among them, thus unifying the whole framework.

\noindent{\textbf{Feature Interaction.}} We combine the density estimation and density decoupling into an unified framework with a shared backbone in an end-to-end manner.
Through a joint optimization, this encourages the co-evolution of backbone features by using sharing weights for different tasks and reduces the parameters of the network. 

\noindent{\textbf{Scale Interaction.}} It is widely acknowledged that high resolution features with more detailed textures are useful to detect tiny objects while low resolution features with more contextual information are useful to suppress false alarms. However, for crowd counting, features with rich contextual information are also demonstrated useful for the estimation in highly congested regions \cite{Guo2019}.

Therefore, in order to make full use of multi-scale features $ \{\mathbf{X}_i\}_{i=1}^{n}$ adaptively, we introduce the Scale Adaptive Feature Fusion (SAFF) block. 
Inspired by SENet \cite{Hu2020}, we use a learnable channel-wise parameter, $\mathbf{w}$, to multiply with each transformed feature, through which it can learn to selectively emphasise informative features and suppress less useful ones from other layers. The output feature $\mathbf{\hat X}$ is formulated as
\begin{equation}
\mathbf{\hat{X}}_{i} = \mathbf{X}_{i} + \sum_{k=1}^n{\mathbf{w}_{i, k} F_{{i,k}}(\mathbf{X}_{k}}) \mathbbm{1}_{[k \neq i]},
\end{equation}
where $\mathbf{w}_{i, k} \in \mathbb{R}^{C}$ is a channel-wise learnable parameter, $C$ is the number of channel of $\mathbf{X}_k$, and $F$ is an up-sampling or down-sampling operation according to the sizes of $\mathbf{X}_i$ and $\mathbf{X}_k$. 
The up-sampling operation uses a set of 1$\times$1 convolutional blocks and bilinear up-sampling layers, and the down-sampling operation uses a set of 3$\times$3 convolutional blocks with a stride 2. $\mathbbm{1}_{[k \neq i]}\in\{0,1\}$ is an indicator function evaluating to 1 iff $k\neq{i}$.  The transformed feature pyramid is in the same size with input features but much richer contexts than the original.

\noindent{\textbf{Gradient Interaction.}}
For the decoupling branch, it would be helpful to its learning if the density intensity in certain regions is aware of. While for the regression branch, it could pay less attention to regions which are classified as background in the decoupling branch. So we construct gradient interaction by multiplying soft masks $\mathbf{\hat M}$ into intermediate density maps $\mathbf{D}$, resulting in FDEM and DDM optimized jointly by the two losses simultaneously.

This helps to learn density value regression and density-specific classification jointly so that the density-aware knowledge contained in the regression task can be leveraged by the classification task. At the same time, by a joint supervision from the regression loss in FDEM, the density-aware knowledge could also be transferred to classification tasks. 

\section{Experiments}

In this section, we firstly describe the experimental settings about datasets, evaluation metrics, and implementation details. Then the effectiveness of the proposed DDM and FDEM is evaluated on the benchmark datasets. Finally, the performance of HDNet and the comparisons with state-of-the-art crowd counting estimators are presented.
\vspace{-0.6em}
\subsection{Experimental setups}
\noindent{\textbf{Datasets.}}
We evaluate the HDNet on four most challenging datasets: ShanghaiTech PartA dataset \cite{Zhang2016}, UCF-QNRF \cite{Idrees2018}, JHU-CROWD++ \cite{Sindagi2020}, and NWPU-Crowd \cite{Wang2021}.


\noindent{\textbf{Evaluation metrics.}}
As in previous works \cite{Zhang2016}, we adopt the Mean Absolute Error (MAE) and the Mean Squared Error (MSE) to evaluate our method. The MAE and MSE are defined by
\begin{equation}
MAE=\frac{1}{N} \sum_{i=1}^{N}\left|\mathbf{D}_{i}^{gt}-\mathbf{\widetilde{D}}_{i}\right|,
\end{equation}
\begin{equation}
MSE=\sqrt{\frac{1}{N}\sum_{i=1}^{N}\left|\mathbf{D}_{i}^{gt}-\mathbf{\widetilde{D}}_{i}\right|^{2}},
\end{equation}
where $N$ is the number of the test images, $\mathbf{D}_{i}^{gt}$ and $\mathbf{\widetilde{D}}_{i}$ are the ground-truth and estimated counts of the $i^{th}$ image, respectively.

\noindent{\textbf{Training.}}
We adopt HRNet \cite{Wang2020HRnet} as the backbone and set $\lambda=1$ to balances two losses in Eq.~\ref{loss}. SGD is used to optimize the model with the learning rate of 0.001, and the weight decay is set as 0.0005. The training batch size is 6. We resize the images to ensure that the longer side is 2048 for all datasets. The number of density levels is set to 3 in the experiments. 

\noindent{\textbf{Ground Truth Generation.}}
Ground truth annotations for crowd counting typically consist of a set of coordinates that indicate the center point of the human head. We follow the standard procedure of the generation of ground truth \cite{Wang2021}, which converts the points to crowd density map using a Gaussian kernel with standard deviation of 15 pixels.
\vspace{-0.6em}
\subsection{Ablation Study}
In this section, we perform ablation studies on ShanghaiTech PartA dataset to analyze effectiveness of proposed modules. 

\begin{figure}[t]
\centering
\includegraphics[width=1.0\columnwidth]{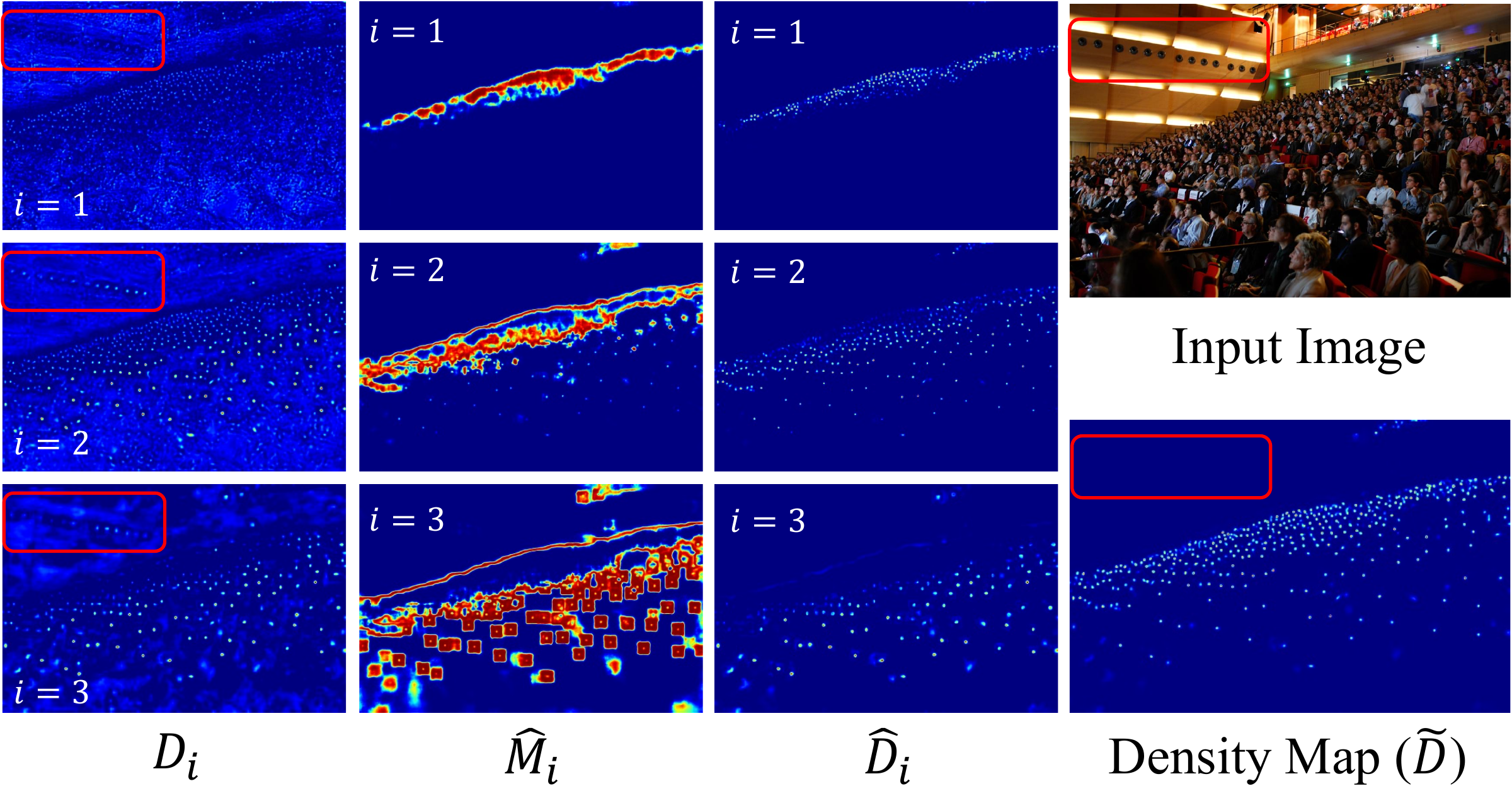} 
\caption{Examples of initial density maps $\mathbf{D}$, soft masks $\mathbf{\hat M}$, intermediate density maps $\mathbf{\hat D}$ and final density map $\mathbf{\widetilde{D}}$.}
\label{Fig3_EaxmpleOfDensityMap}
\end{figure}

\begin{table}[htbp]
    \fontsize{9}{10}\selectfont
    \centering{
    \begin{tabular}{c|c|rr}
        \toprule[1pt]
        \textbf{FB} Decoupling & \textbf{FD} Decoupling & MAE & MSE \\
        \midrule[0.5pt]
        $\times$&$\times$ & 59.42 & 102.71 \\
        $\surd$&$\times$  & 56.98 & 94.01 \\
        $\surd$&$\surd$  & \textbf{54.61} & \textbf{92.13} \\
        \bottomrule[1pt]
    \end{tabular}}
    \vspace{-5pt}
    \caption{Ablation study of decoupling strategies.}
    \label{tab1_Decoupling}
\end{table}

\noindent{\textbf{Foreground and Background Decoupling.}}
To solve the cluttered background noise, we adopt the decoupling foreground and background method. 
We define a base estimator only with one density head as the baseline.
As shown in first row of Table \ref{tab1_Decoupling}, the baseline achieves an MAE of 59.42. While equipped with FB Decoupling, HDNet ($n=1$, w/o SAFF) achieves 56.98 MAE. There is no doubt that it shows the importance of decoupling foreground and background strategy.

\begin{table}[htbp]
    \fontsize{9}{10}\selectfont
    \centering{
    \begin{tabular}{c|rr}
        \toprule[1pt]
        Feature Interaction & MAE & MSE \\
        \midrule[0.5pt]
        $\times$ & 55.77  & 100.98 \\
        $\surd$ & \textbf{54.61} & \textbf{92.13} \\
        \bottomrule[1pt]
    \end{tabular}}
     \vspace{-5pt}
    \caption{Ablation study of feature interaction.}
    \label{tab2_BackboneFeatureSharing}
\end{table}
\begin{table}[htbp]
    \fontsize{9}{10}\selectfont
    \centering{
    \begin{tabular}{c|c|rr}
        \toprule[1pt]
        SAFF & channel-wise parameter ($\mathbf{w}$) & MAE & MSE \\
        \midrule[0.5pt]
        $\times$ & - & 54.61 & 92.13 \\
        $\surd$ & $\times$ & 54.60 & 93.63 \\
        $\surd$ & $\surd$ & \textbf{53.39} & \textbf{89.87} \\
        \bottomrule[1pt]
    \end{tabular}}
    \vspace{-5pt}
    \caption{Ablation study of SAFF. }
    \label{tab4_ScaleInteraction}
\end{table}

\begin{table}[htbp]
    \fontsize{9}{9}\selectfont
    \centering{
    \begin{tabular}{c|rr}
        \toprule[1pt]
        Gradient Interaction & MAE & MSE \\
        \midrule[0.5pt]
        $\times$ & 56.60 & 95.71 \\
        $\surd$ & \textbf{53.39} & \textbf{89.87} \\
        \bottomrule[1pt]
    \end{tabular}}
    \vspace{-5pt}
    \caption{Ablation study of gradient interaction.}
    \label{tab3_GradientInteraction}
\end{table}

\noindent{\textbf{Foreground Density Decoupling.}}
To demonstrate the effectiveness of decoupling foreground density, the foreground density decoupling further divides the foreground into multiple density levels according to its density. 
We define the baseline with Foreground and Background (\textbf{FB}) Decoupling and Foreground Density (\textbf{FD}) Decoupling as HDNet ($n=3$, w/o SAFF). As shown in Table \ref{tab1_Decoupling},  it obtains an MAE of 54.61. 
Compared  HDNet ($n=1$, w/o SAFF), the MAE decreases by 4.2\%, which proves the necessity of decoupling foreground density decoupling. 
FB Decoupling ignores the large intra-class variance within foreground regions, while FD Decoupling provides a fine-grained supervision. It helps the accurate modeling for background region owing to the decreasing intra-class variance. At the same time, FD Decoupling allows each task-specific expert to focus on their most skilled sub-task, thus reducing the risk of overfitting. 

Fig.\ \ref{Fig3_EaxmpleOfDensityMap} shows examples of initial density maps $\mathbf{D}$, soft masks $\mathbf{\hat M}$, intermediate density maps $\mathbf{\hat D}$ and final density map $\mathbf{\widetilde{D}}$. Density Decoupling and Interaction suppress noise in red rectangles successfully, which also make each expert of density estimation focus on the density-specific region.

 \begin{table}[t]
\setlength\tabcolsep{3pt}
\fontsize{7}{9}\selectfont
\centering
\begin{tabular}{l|c|rr|rr|rr|rr} 
\toprule
\multirow{2}{*}{Methods} &\multirow{2}{*}{Venue} & \multicolumn{2}{c|}{NWPU}& \multicolumn{2}{c|}{UCF-QNRF}& \multicolumn{2}{c|}{JHU++}& \multicolumn{2}{c}{ShTechA}   \\
\cline{3-10}  
   & &   MAE & MSE&    MAE & MSE&    MAE & MSE&    MAE & MSE \\
  \midrule
CSRNet \cite{Li2018} & CVPR & 121.3 & 387.8  & 110.6 & 190.1& 85.9  & 309.2 & 68.2  & 115.0 \\
SANet \cite{cao2018scale} & ECCV & 190.6 & 491.4 & -     & -   & 91.1  & 320.4   & 67.0    & 104.5 \\
DSSINet \cite{Liu2019}& ICCV  & -     & -      & 99.1  & 159.2 & 133.5 & 416.5& 60.6  & 96.0 \\
MBTTBF  \cite{Sindagi2019}& ICCV& -  & - & 97.5  & 165.2& 81.8  & 299.1 & 60.2  & 94.1 \\
BL \cite{Ma2019}& ICCV & 105.4 & 454.2  & 88.7  & 154.8 & 75.0  & 299.9& 62.8  & 101.8 \\
LSCCNN \cite{BabuSam2020}& TPAMI& -    & -& 120.5 & 218.2 & 112.7 & 454.4& 66.5  & 101.8 \\
KDMG \cite{wan2020kernel}& TPAMI & 100.5 & 415.5  & 99.5  & 173.0& 69.7  & 268.3  & 63.8  & 99.2 \\
ASNet \cite{Jiang2020}& CVPR & -   & -  & 91.6  & 159.7 & -   & -    & 57.8  & 90.1 \\
LibraNet \cite{Liu2020}& ECCV  & - & - & 88.1  & \underline{143.7} & - & - & 55.9  & 97.1 \\
DM-count \cite{Wang2020}& NIPS  & 88.4  & 357.6  & 85.6  & 148.3& 68.4  & 283.3 & 59.7  & 95.7 \\
NoiseCC \cite{Wan2020}& NIPS  & 96.9  & 534.2  & 85.8  & 150.6 & 67.7  & \underline{258.5}& 61.9  & 99.6 \\
LA-Batch \cite{Zhou2021}& TPAMI & -  & -   & 113.0  & 210.0 & -  &-& 65.8  & 103.6 \\
UOT \cite{Ma2021} & AAAI& 87.8  & 387.5  & \underline{83.3}  & \textbf{142.3}&  \underline{60.5}  & \textbf{252.7} &58.1& 95.9 \\
SASNet \cite{song2021choose} & AAAI& -  & -   & 85.2  & 147.3& -  & - & \underline{53.6}  & \textbf{88.4}\\
GLoss \cite{JiaWan2021} & CVPR & \underline{79.3}  & \underline{346.1}  & 84.3  & 147.5 & \textbf{59.9}  & 259.5& 61.3  & 95.4 \\
HDNet (Ours)  & -& \textbf{76.1} & \textbf{322.3}  & \textbf{83.2} & 148.3 & 62.9 & 276.2& \textbf{53.4} & \underline{89.9} \\
  \bottomrule
  \end{tabular}
  \caption{Comparison with state-of-the-art methods on four challenging datasets. Smaller number indicates better performance. In each column, the best result is \textbf{blod}, and the second best is \underline{underlined.}}
  \label{tab6_ComparedWithSOTA}
\end{table}

\noindent{\textbf{Feature Interaction.}}
In order to verify the effectiveness of Feature Interaction, we split the HDNet ($n=3$, w/o SAFF) into two independent networks (i.e., FDEM and DDM). Here, each network has its own backbone.
Compared to the HDNet with a sharing backbone, the two independent networks without Feature Interaction increase an MAE from 54.61 to 55.77, as shown in Table \ref{tab2_BackboneFeatureSharing}. This proves that the feature sharing of the two tasks can encourage the co-evolution of the backbone feature and achieve a better performance.

\noindent{\textbf{Scale Interaction.}}
Follow the setting in the previous paragraph, the HDNet ($n=3$) without SAFF block obtains an MAE of 54.61, as shown in first row of Table \ref{tab4_ScaleInteraction}. The SAFF without the channel-wise parameter, $\mathbf{w}$, brings a slight performance improvement on MAE but a larger MSE of 93.63. The reason could be that the complete fusion of different resolution features results in feature redundancy. In the last row, a channel-wise parameter $\mathbf{w}$ is introduced in SAFF. The $\mathbf{w}$ selectively emphasises informative features and suppresses less useful ones from other layers, achieving an MAE of 53.39 and an MSE of 89.87. This improved performance shows that Scale Interaction can enable the network to adaptively fuse complementary features, resulting in increasing the density-aware ability of different density regions.

\noindent{\textbf{Gradient Interaction.}}
To illustrate the improvement of the gradient interaction, we change soft masks $\mathbf{\hat M}$ in the DDM to truncated binary masks, which cuts off the back propagation between FDEM and DDM in the training stage.
As shown in Table \ref{tab3_GradientInteraction}, this change obtains an MAE of 56.60, while using soft masks with back propagation ability can get an MAE of 53.39. It shows that Gradient Interaction ensures a reciprocal optimization between FDEM and DDM by propagating gradient signals to each other. This leads to a better knowledge sharing among tasks and improve each other.

\begin{figure}[t]
\centering
\includegraphics[width=1.0\columnwidth]{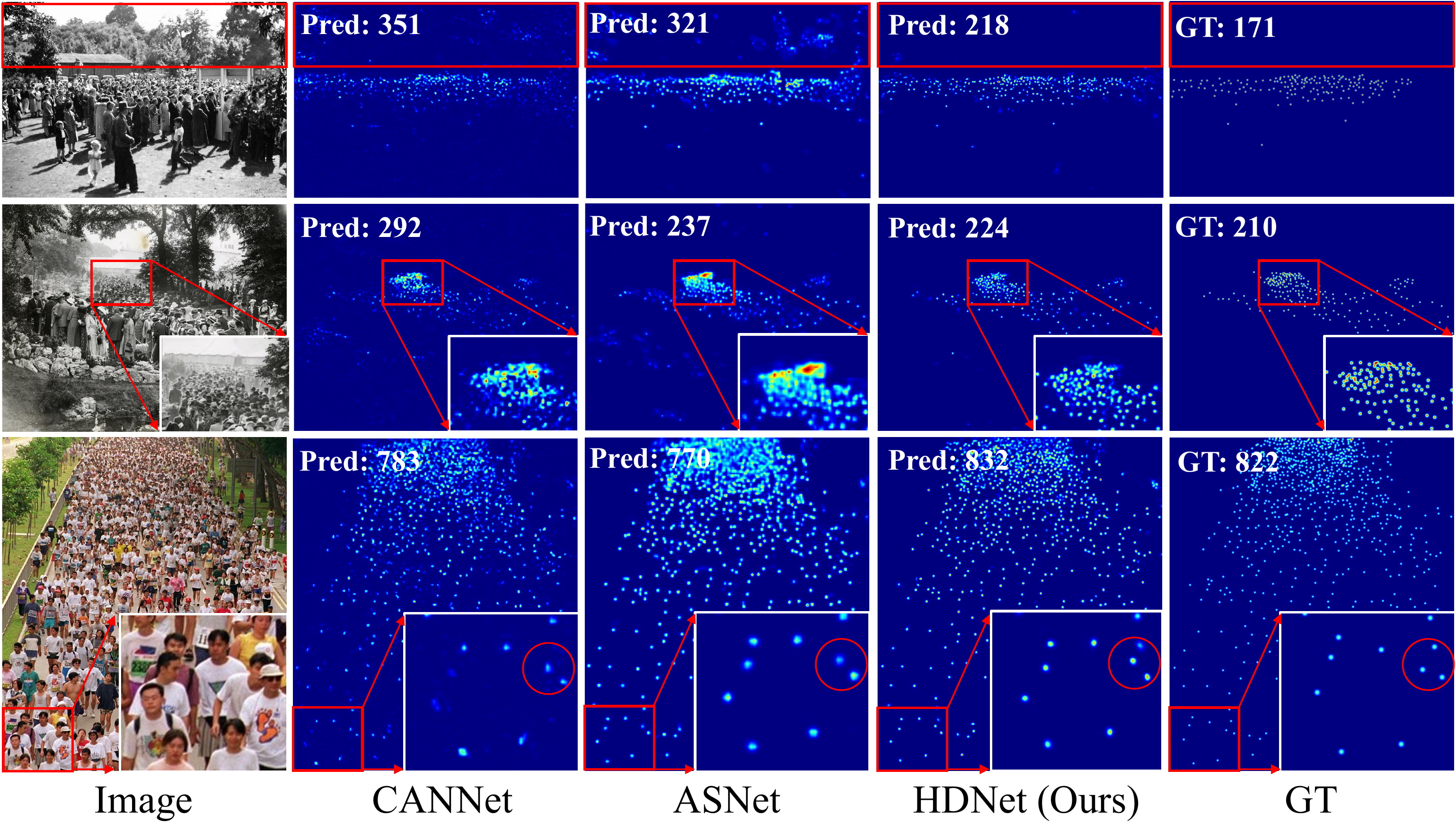} 
\caption{The demonstration of different methods on the ShanghaiTech PartA dataset. GT means the ground truth.}
\label{fig4}
\end{figure}

\subsection{Comparisons with State-of-the-Arts}
To demonstrate the effectiveness of our proposed approach, we compare our approach with state-of-the-art methods on four challenging datasets with various densities. The results are illustrated in Table \ref{tab6_ComparedWithSOTA}. As we can see that our proposed approach is the state-of-the-art or close to state-of-the-art on the four challenging datasets.
 
Fig. \ref{fig4} compares the density maps of different methods from left to right columns on the ShanghaiTech PartA dataset. Compared with CANNet \cite{Liu2019} (second column) and ASNet \cite{Jiang2020} (third column), the proposed HDNet significantly eliminates noises caused by cluttered backgrounds. For the second row, the dense crowd at the center, our method obtains a more discrete density prediction due to the use of the scale adaptive features with high-resolution and more context. For the last row, HDNet reserves more consistency with the real crowd distributions and is robust to occlusions. It shows that HDNet is very powerful and achieves much more accurate count estimations.

\section{Conclusion}

In this work, we firstly propose a novel hierarchically decoupled strategy to simultaneously solve two long-standing problems: the cluttered background noise and the large density variation. Specifically, a Density Decoupling Module is proposed to guide this key decoupling process, which is supervised by a fine-grained density-aware learning target. Then the decoupled components are distributed to several task-specific experts according to their most skilled sub-task. As a complement to this effective decoupling strategy, three kinds of interaction strategies are proposed to collaboratively integrate those decoupled components. By combining these spirits together, we propose a compact, effective and unified counting model named as HDNet. The effectiveness of our contributions is demonstrated by the state-of-the-art performance on several dominant counting benchmarks.


\bibliographystyle{IEEEbib}
\bibliography{icme2022template}

\begin{thebibliography}{10}

\bibitem{Modolo2021}
Davide Modolo, Bing Shuai, Rahul~Rama Varior, and Joseph Tighe,
\newblock ``{Understanding the impact of mistakes on background regions in
  crowd counting},''
\newblock in {\em WACV}, 2021.

\bibitem{song2021choose}
Qingyu Song, Changan Wang, Yabiao Wang, Ying Tai, Chengjie Wang, Jilin Li, Jian
  Wu, and Jiayi Ma,
\newblock ``{To Choose or to Fuse? Scale Selection for Crowd Counting},''
\newblock in {\em AAAI}, 2021.

\bibitem{miao2020shallow}
Yunqi Miao, Zijia Lin, Guiguang Ding, and Jungong Han,
\newblock ``Shallow feature based dense attention network for crowd counting,''
\newblock in {\em AAAI}, 2020.

\bibitem{arteta2016counting}
Carlos Arteta, Victor Lempitsky, and Andrew Zisserman,
\newblock ``{Counting in the wild},''
\newblock in {\em ECCV}, 2016.

\bibitem{wan2019residual}
Jia Wan, Wenhan Luo, Baoyuan Wu, Antoni~B Chan, and Wei Liu,
\newblock ``{Residual regression with semantic prior for crowd counting},''
\newblock in {\em CVPR}, 2019.

\bibitem{Zhang2016}
Yingying Zhang, Desen Zhou, Siqin Chen, Shenghua Gao, and Yi~Ma,
\newblock ``{Single-image crowd counting via multi-column convolutional neural
  network},''
\newblock in {\em CVPR}, 2016.

\bibitem{Li2018}
Yuhong Li, Xiaofan Zhang, and Deming Chen,
\newblock ``{CSRNet: Dilated Convolutional Neural Networks for Understanding
  the Highly Congested Scenes},''
\newblock in {\em CVPR}, 2018.

\bibitem{Guo2019}
Dan Guo, Kun Li, Zheng~Jun Zha, and Meng Wang,
\newblock ``{DadNet: Dilated-attention-deformable convnet for crowd
  counting},''
\newblock in {\em ACM MM}, 2019.

\bibitem{liu2019adcrowdnet}
Ning Liu, Yongchao Long, Changqing Zou, Qun Niu, Li~Pan, and Hefeng Wu,
\newblock ``{Adcrowdnet: An attention-injective deformable convolutional
  network for crowd understanding},''
\newblock in {\em CVPR}, 2019.

\bibitem{Jiang2020}
Xiaoheng Jiang, Li~Zhang, Mingliang Xu, Tianzhu Zhang, Pei Lv, Bing Zhou, Xin
  Yang, and Yanwei Pang,
\newblock ``{Attention scaling for crowd counting},''
\newblock in {\em CVPR}, 2020.

\bibitem{Hossain2019}
Mohammad~Asiful Hossain, Mehrdad Hosseinzadeh, Omit Chanda, and Yang Wang,
\newblock ``{Crowd counting using scale-aware attention networks},''
\newblock in {\em WACV}, 2019.

\bibitem{Hu2020}
Jie Hu, Li~Shen, Samuel Albanie, Gang Sun, and Enhua Wu,
\newblock ``{Squeeze-and-Excitation Networks},''
\newblock {\em TPAMI}, 2020.

\bibitem{Idrees2018}
Haroon Idrees, Muhmmad Tayyab, Kishan Athrey, Dong Zhang, Somaya Al-Maadeed,
  Nasir Rajpoot, and Mubarak Shah,
\newblock ``{Composition Loss for Counting, Density Map Estimation and
  Localization in Dense Crowds},''
\newblock in {\em ECCV}, 2018.

\bibitem{Sindagi2020}
Vishwanath Sindagi, Rajeev Yasarla, and Vishal~M.M. Patel,
\newblock ``{JHU-CROWD++: Large-Scale Crowd Counting Dataset and A Benchmark
  Method},''
\newblock {\em TPAMI}, 2020.

\bibitem{Wang2021}
Qi~Wang, Junyu Gao, Wei Lin, and Xuelong Li,
\newblock ``{NWPU-Crowd: A Large-Scale Benchmark for Crowd Counting and
  Localization},''
\newblock {\em TPAMI}, 2020.

\bibitem{Wang2020HRnet}
Jingdong Wang, Ke~Sun, Tianheng Cheng, Borui Jiang, Chaorui Deng, Yang Zhao,
  Dong Liu, Yadong Mu, Mingkui Tan, Xinggang Wang, Wenyu Liu, and Bin Xiao,
\newblock ``{Deep High-Resolution Representation Learning for Visual
  Recognition},''
\newblock {\em TPAMI}, 2020.

\bibitem{cao2018scale}
Xinkun Cao, Zhipeng Wang, Yanyun Zhao, and Fei Su,
\newblock ``{Scale aggregation network for accurate and efficient crowd
  counting},''
\newblock in {\em ECCV}, 2018.

\bibitem{Liu2019}
Lingbo Liu, Zhilin Qiu, Guanbin Li, Shufan Liu, Wanli Ouyang, and Liang Lin,
\newblock ``{Crowd counting with deep structured scale integration network},''
\newblock in {\em ICCV}, 2019.

\bibitem{Sindagi2019}
Vishwanath Sindagi and Vishal Patel,
\newblock ``{Multi-level bottom-top and top-bottom feature fusion for crowd
  counting},''
\newblock in {\em ICCV}, 2019.

\bibitem{Ma2019}
Zhiheng Ma, Xing Wei, Xiaopeng Hong, and Yihong Gong,
\newblock ``{Bayesian loss for crowd count estimation with point
  supervision},''
\newblock in {\em ICCV}, 2019.

\bibitem{BabuSam2020}
Deepak {Babu Sam}, Skand~Vishwanath Peri, Mukuntha {Narayanan Sundararaman},
  Amogh Kamath, and Venkatesh~Babu Radhakrishnan,
\newblock ``{Locate, Size and Count: Accurately Resolving People in Dense
  Crowds via Detection},''
\newblock {\em TPAMI}, 2020.

\bibitem{wan2020kernel}
Jia Wan, Qingzhong Wang, and Antoni~B Chan,
\newblock ``{Kernel-based density map generation for dense object counting},''
\newblock {\em TPAMI}, 2020.

\bibitem{Liu2020}
Liang Liu, Hao Lu, Hongwei Zou, Haipeng Xiong, Zhiguo Cao, and Chunhua Shen,
\newblock ``{Weighing Counts: Sequential Crowd Counting by Reinforcement
  Learning},''
\newblock in {\em ECCV}, 2020.

\bibitem{Wang2020}
Boyu Wang, Huidong Liu, Dimitris Samaras, and Minh Hoai,
\newblock ``{Distribution matching for crowd counting},''
\newblock in {\em NeurIPS}, 2020.

\bibitem{Wan2020}
Jia Wan and Antoni Chan,
\newblock ``{Modeling Noisy Annotations for Crowd Counting},''
\newblock in {\em NeurIPS}, 2020.

\bibitem{Zhou2021}
Joey~Tianyi Zhou, Le~Zhang, Du~Jiawei, Xi~Peng, Zhiwen Fang, Zhe Xiao, and
  Hongyuan Zhu,
\newblock ``{Locality-Aware Crowd Counting},''
\newblock {\em TPAMI}, 2021.

\bibitem{Ma2021}
Zhiheng Ma, Xing Wei, Xiaopeng Hong, Hui Lin, Yunfeng Qiu, and Yihong Gong,
\newblock ``{Learning to Count via Unbalanced Optimal Transport},''
\newblock in {\em AAAI}, 2021.

\bibitem{JiaWan2021}
{Jia Wan},
\newblock ``{A Generalized Loss Function for Crowd Counting and
  Localization},''
\newblock in {\em CVPR}, 2021.

\end{thebibliography}

\end{document}